# Algorithm Configurations of MOEA/D with an Unbounded External Archive

Lie Meng Pang, Hisao Ishibuchi*, Ke Shang
Guangdong Provincial Key Laboratory of Brain-inspired Intelligent Computation
Department of Computer Science and Engineering, Southern University of Science and Technology
Shenzhen, China
Email: panglm@sustech.edu.cn, hisao@sustech.edu.cn, kshang@foxmail.com

*Abstract*— In the evolutionary multi-objective optimization (EMO) community, it is usually assumed that the final population is presented to the decision maker as the result of the execution of an EMO algorithm. Recently, an unbounded external archive was used to evaluate the performance of EMO algorithms in some studies where a pre-specified number of solutions are selected from all the examined non-dominated solutions. In this framework, which is referred to as the solution selection framework, the final population does not have to be a good solution set. Thus, the solution selection framework offers higher flexibility to the design of EMO algorithms than the final population framework. In this paper, we examine the design of MOEA/D under these two frameworks. First, we show that the performance of MOEA/D is improved by linearly changing the reference point specification during its execution through computational experiments with various combinations of initial and final specifications. Robust and high performance of the solution selection framework is observed. Then, we examine the use of a genetic algorithm-based offline hyper-heuristic method to find the best configuration of MOEA/D in each framework. Finally, we further discuss solution selection after the execution of an EMO algorithm in the solution selection framework.

*Keywords*— Algorithm configurations, evolutionary multi-objective optimization (EMO), final population framework, solution selection framework

I. INTRODUCTION

The development of effective and efficient algorithms to solve multi-objective optimization problems (MOPs) has been an active research topic in the evolutionary computation (EC) community for more than three decades [1]. The conflicting nature among the objectives of an MOP makes it almost impossible to obtain a single optimal solution for all objectives. Instead, a set of trade-off solutions (also known as Pareto optimal solutions) is usually obtained. These Pareto optimal solutions form the Pareto front in the objective space. Evolutionary multi-objective optimization (EMO) algorithms have shown their usefulness in finding a set of non-dominated solutions to approximate the Pareto front in a single run. For this reason, various EMO algorithms have been proposed to solve MOPs. These EMO algorithms are designed based on different types of fitness evaluation mechanisms, such as Pareto dominance-based [2], [3], decomposition-based [4], [5] and indicator-based [6], [7] mechanisms. The common goal of these EMO algorithms is to obtain a high quality (in terms of convergence, diversity and spread) final population to approximate the Pareto front.

One issue for almost all EMO algorithms is that they may lose some good solutions along the evolutionary process. In order to keep the good solutions found during the evolution, elitism mechanisms are used. The elitism mechanism is usually accomplished by using the ($\mu + \lambda$)-selection strategy, where $\mu$ is the number of parents (i.e., solutions in the current population) and $\lambda$ is the number of offspring. In the ($\mu + \lambda$)-selection strategy, $\mu$ parents compete with $\lambda$ offspring in a combined population. Then, the best $\mu$ solutions are selected from the combined population. The selected $\mu$ solutions are solutions (parents) in the next generation. NSGA-II [2] is one of the most popular EMO algorithms that use the ($\mu + \lambda$)-selection strategy as an elitism mechanism.

Another kind of elitism mechanism is the use of an external archive [8]. The external archive (also known as the secondary population) is responsible for storing non-dominated solutions found during the evolution process. For computational efficiency, a bounded external archive (i.e., an archive with a pre-specified capacity) is often used. Some EMO algorithms such as SPEA2 [3] and PAES [9] allow solutions in the external archive to participate in the evolutionary process. Other EMO algorithms simply return the solutions stored in the archive as the final result of an algorithm run. However, since the bounded external archive can only store a certain number of solutions, its optimality cannot be guaranteed. That is, the monotonicity property of the bounded external archive may not be fulfilled. This is because some solutions must be discarded when the archive exceeds the pre-specified capacity. Since the archive management is performed at each generation, it is not necessarily optimal in the long run. As a result, some solutions in the final archive can be dominated by discarded solutions in a previous generation [10]. This problem can be resolved by using an unbounded external archive. One may think that maintaining an unbounded external archive requires a huge computational cost. However, as pointed out in [11], the storing cost for an unbounded external archive is much lower (or even negligible) if it is compared with the expensive function evaluation cost of real-world problems.

This work was supported by National Natural Science Foundation of China (Grant No. 61876075), Guangdong Provincial Key Laboratory (Grant No. 2020B121201001), the Program for Guangdong Introducing Innovative and Enterpreneurial Teams (Grant No. 2017ZT07X386), Shenzhen Science and Technology Program (Grant No. KQTD2016112514355531), the Program for University Key Laboratory of Guangdong Province (Grant No. 2017KSYS008).
*Corresponding Author: Hisao Ishibuchi (hisao@sustech.edu.cn)

Recently, an unbounded external archive was used to evaluate the performance of EMO algorithms in some studies [11], [12], where a pre-specified number of solutions are selected from all the examined non-dominated solutions. Such a framework, which is referred to as the solution selection framework in this paper, can also be incorporated in the algorithm design process. In the solution selection framework, the final population does not have to be a good solution set. Instead, a set of good solutions to be presented to the decision makers is selected from the unbounded external archive. Thus, the solution selection framework offers higher flexibility to the design of EMO algorithms than the existing final population framework (and the bounded archive framework).

This paper studies the effect of incorporating the solution selection framework into the algorithm design of EMO algorithms. The multi-objective evolutionary algorithm based on decomposition (MOEA/D) [5] is used to demonstrate the flexibility offered by the solution selection framework. MOEA/D is one of the most popular EMO algorithms. It has been widely used for its high search ability, high computation efficiency, and high scalability to many-objective optimization problems [13]. In this paper, we show that the performance of MOEA/D is improved by linearly changing the reference point specification during its execution. Various combinations of initial and final specifications of the reference point are examined using the final population framework and the solution selection framework. The WFG1-4 [14] and Minus-WFG1-4 [15] test problems are used in our experiments. Experimental results suggest that MOEA/D with the solution selection framework has more robust and higher performance than that with the final population framework.

In order to further examine the robustness and flexibility of the solution selection framework for the design of MOEA/D, a genetic algorithm (GA) is used as an offline hyper-heuristic method to find the best configuration of MOEA/D in each framework. The GA-based hyper-heuristic method selects a suitable scalarizing function as well as the initial and final reference point specifications for each test problem. We show that the search for the optimal algorithm configuration in the final population framework is not easy for a GA-based hyper-heuristic due to the stochastic search nature of MOEA/D. However, for the solution selection framework, good performance of MOEA/D can be easily obtained by the GA-based hyper-heuristic method.

The organization of this paper is as follows. First, Section II gives the background of MOP and MOEA/D. Next, Section III presents experimental results by MOEA/D with each framework. Results of a GA-based hyper-heuristic method are also reported in Section III. Finally, Section IV concludes the paper.

## II. BACKGROUND

### A. Multi-Objective Optimization Problems

Without loss of generality, we consider minimization problems in this paper. Thus, an MOP is written as:

$$\text{Minimize } f_1(x), f_2(x), \dots, f_M(x) \text{ subject to } x \in X, \quad (1)$$

where $x = (x_1, \dots, x_D)$ is a $D$-dimensional decision vector, $X$ is the feasible region of $x$, and $f_i(x)$ is the $i$-th objective to be minimized ($i = 1, 2, \dots, M$).

A solution $x^A$ is said to Pareto dominate $x^B$ iff $f_i(x^A) \leq f_i(x^B)$ for all $i \in \{1,2,\dots,M\}$ and $f_j(x^A) < f_j(x^B)$ for at least one index $j \in \{1,2,\dots,M\}$. $x^*$ is a Pareto optimal solution if it is not dominated by any other solution in $X$. The Pareto optimal set (PS) consists of all Pareto optimal solutions, and the projection of the PS to the objective space forms the Pareto front (PF).

### B. MOEA/D

MOEA/D decomposes an MOP with $M$ objectives into $N$ single-objective sub-problems using a set of weight vectors $W = \{w^1, w^2, \dots, w^N\}$ and a scalarizing function $g$. Each weight vector $w^i = (w_1^i, w_2^i, \dots, w_M^i)^T$ must fulfil the following relation:

$$\sum_{j=1}^{M} w_j^i = 1 \text{ and } w_j^i \geq 0 \ (j = 1,2,\dots,M), \quad (2)$$

where $i \in \{1,2,\dots,N\}$. In our study, the Das and Dennis method [16] is used to systematically generate the weight vectors. Each subproblem $i$ has a single individual $x^i$ ($i \in \{1,2,\dots,N\}$). Thus, the population size is equal to the number of sub-problems (which is also equal to the number of the generated weight vectors). MOEA/D uses a scalarizing function to calculate the fitness value of each individual. Each weight vector has a neighborhood which is defined by the Euclidean distance (the weight vector itself is also included in its own neighborhood). The environmental selection and the replacement of individuals are performed within the neighborhood. A new solution (i.e., offspring) is generated from the parents in the neighborhood. The generated offspring is compared with each solution in the neighborhood. If the offspring is better with respect to the weight vector of the current solution, the current solution is replaced by the offspring.

A scalarizing function plays an important role in MOEA/D. Since different scalarizing functions have different search behavior [17], the choice of an appropriate scalarizing function strongly depends on the characteristics of the problem at hand. In this paper, the four commonly used scalarizing functions are considered. They are the weighted sum ($g^{\text{WS}}$), Tchebycheff ($g^{\text{TCH}}$), modified Tchebycheff ($g^{\text{MTCH}}$), and Penalty-based Boundary Intersection (PBI) ($g^{\text{PBI}}$) functions. The four scalarizing functions are defined as follows:

$$\text{Minimize } g^{\text{WS}}(x|w) = w_1 f_1(x) + \cdots + w_M f_M(x), \quad (3)$$

$$\text{Minimize } g^{\text{TCH}}(x|w, z^*) = \max_{i=1,2,\dots,M} \{w_i \cdot |z_i^* - f_i(x)|\}, \quad (4)$$

$$\text{Minimize } g^{\text{MTCH}}(x|w, z^*) = \max_{i=1,2,\dots,M} \{|z_i^* - f_i(x)|/w_i\}, \quad (5)$$

$$\text{Minimize } g^{\text{PBI}}(x|w, z^*) = d_1 + \theta d_2, \quad (6)$$

$$d_1 = |(f(x) - z^*)^T w|/\|w\|, \quad (7)$$

$$d_2 = \|f(x) - z^* - d_1(w/\|w\|)\|, \quad (8)$$

where $z^* = (z_1^*, z_2^*, \dots, z_M^*)$ is the reference point for the Tchebycheff, modified Tchebycheff, and PBI functions. The ideal point is used to specify the reference point. However, the true ideal point is usually unknown for real-world problems. The common practice is to estimate each element $z_i^*$ ($i \in \{1,2,...,M\}$) of $\mathbf{z}^*$ by the minimum value of $f_i(\mathbf{x})$ among all solutions examined so far. For the modified Tchebycheff function in (5), if $w_i = 0$, $w_i$ is set to $10^{-6}$ to avoid division by zero. For the PBI function in (6), the penalty parameter $\theta$ is a user-definable non-negative real number.

A simple normalization mechanism has been included in the original MOEA/D paper [5] to deal with problems with disparately scaled objectives. The necessity of normalization in MOEA/D has been discussed in [18]. In this paper, the normalization mechanism in [18] is used. The estimated ideal point $\mathbf{z}^*$ and the nadir point $\mathbf{z}^{nad}$ are used to normalize the objective value $z_i$ of the $i$th-objective $f_i(\mathbf{x})$ as follows:

$$z_i := \frac{z_i - z_i^*}{z_i^{nad} - z_i^* + \varepsilon}, \tag{9}$$

where $\varepsilon$ is a small non-negative number used to prevent the denominator from becoming zero when $z_i^{nad} = z_i^*$. In this paper, $\varepsilon$ is set to $10^{-6}$. Each element in $\mathbf{z}^{nad} = (z_1^{nad}, z_2^{nad}, ..., z_M^{nad})$ is specified by the maximum value of each objective $f_i(\mathbf{x})$ among all solutions in the current population.

*C. Reference Point Specification for MOEA/D*

The effect of the reference point specification for MOEA/D has been discussed in [19], [20]. A reference point in MOEA/D serves as the origin for the weight vectors, which determines the search region of MOEA/D in the objective space. The location of the reference point affects the performance of MOEA/D. As explained in Section II.B, the estimated ideal point is usually used as the reference point. That is, $\mathbf{z}^* = \mathbf{z}^{min}$ where $\mathbf{z}^{min} = (z_1^{min}, z_2^{min}, ..., z_M^{min})$ and $z_i^{min}$ is the minimum value of each objective among all solutions examined so far ($i \in \{1,2,...,M\}$). However, as pointed out in [19], [20], the estimated ideal point in the early generations may not be a proper specification for the reference point. This is because the solutions in the early generations are usually far away from the Pareto front. Thus, the estimated ideal point is likely to mislead the initial phase of evolution. Based on these discussions, it was suggested in [19] to use a much better point (i.e., a point that dominates the estimated ideal point) as the reference point in the initial phase of evolution. Then, as the evolution process progresses, the difference between the reference point and the estimated ideal point is gradually decreased to zero. This can be done using a linearly decreasing mechanism [20] as follows:

$$z_i^* = z_i^{min} - \epsilon_i(t), \tag{10}$$

$$\epsilon_i(t) = (\epsilon_i^{ini} - \epsilon_i^{end})(\frac{T-t}{T-1}) + \epsilon_i^{end}, \tag{11}$$

where $\epsilon_i^{ini}$ and $\epsilon_i^{end}$ are the initial and final settings of $\epsilon_i(t)$, respectively, $T$ is the maximum generation number, and $t$ is the current generation index. In this study, we use the same function of $\epsilon_i(t)$ for all objectives: $\epsilon_1(t) = \epsilon_2(t) = \cdots = \epsilon_M(t)$ in the normalized objective space. Promising results were reported by MOEA/D with the linearly decreasing reference point mechanism [20]. However, it is not easy to appropriately specify the initial and final settings of $\epsilon_i$. This is because their appropriate settings are dependent on the characteristic of each problem.

III. COMPUTATIONAL EXPERIMENTS

*A. Reference Point Specification in MOEA/D*

In this section, we show that an appropriate reference point specification in MOEA/D is important in the final population framework. We also show that high performance is obtained by MOEA/D in a wider range of reference point specifications in the solution selection framework.

In our computational experiments, all combinations of five values $\{-1, 0, 1, 3, 5\}$ are examined for $\epsilon_i^{ini}$ and $\epsilon_i^{end}$ (i.e., 25 combinations in total). During the search process, the solutions are normalized by (9), which changes all objective values within [0, 1]. Thus, $z_i^{min} = 0$. The standard reference point specification for MOEA/D in the literature is $\epsilon_i^{ini} = 0$ and $\epsilon_i^{end} = 0$. That is, the reference point is the same as the estimated ideal point. When $\epsilon_i > 0$, the reference point has a better value than the estimated ideal point. A better value may be useful at the initial stage of the evolutionary process. This is because a wider region in the objective space could be searched using a reference point that is better than the estimated ideal point. Thus, the exploration ability of MOEA/D can be enhanced. When $\epsilon_i < 0$, the reference point is dominated by the estimated ideal point. Such a reference point may help to increase the convergence ability of MOEA/D towards the Pareto front (especially towards the inside of the Pareto front) at the final stage of evolution.

All combinations of the five values for $\epsilon_i^{ini}$ and $\epsilon_i^{end}$ are examined using MOEA/D with different scalarizing functions (i.e., MOEA/D-PBI, MOEA/D-TCH, MOEA/D-MTCH and MOEA/D-WS) on eight test problems: WFG1-4 and Minus-WFG1-4. These test problems were chosen because each of them has different characteristic features and Pareto front shapes. The following experimental settings (which are commonly used in the literature [24]) are used in MOEA/D:

Population size: 91,
Maximum number of solution evaluations: 36,400,
Neighborhood size: 20,
Crossover: Simulated binary crossover with the probability 1 and the distribution index 20,
Mutation: Polynomial mutation with the probability 1/$D$ ($D$ is the number of decision variables) and the distribution index 20.

For MOEA/D-PBI, the penalty parameter $\theta$ is set to 5. For each combination of $\epsilon_i^{ini}$ and $\epsilon_i^{end}$, 31 independent runs of MOEA/D experiments are performed. In the solution selection framework, 91 solutions are selected from the unbounded external archive using the distance-based selection (DSS) method [21]. The inverted generational distance (IGD) indicator is used for evaluating the performance of MOEA/D with different reference point specifications. The smaller IGD value means the better performance. We use the PlatEMO [22] platform for IGD calculation.

Fig. 1 and Fig. 2 show the experimental results on the three-objective WFG2 and Minus-WFG2 test problems, respectively. The number of decision variables is specified as $D = 24$. These figures clearly show that the reference point specification has a large impact on the performance of MOEA/D-PBI, MOEA/D-TCH, and MOEA/D-MTCH. This is because the search regions of these scalarizing functions are determined by the location of the reference point. In principle, the reference point specification has no effect on the performance of MOEA/D with the weighted sum function (since the reference point is not included in the weighted sum formulation). Small differences in the performance of MOEA/D with the weighted sum in each figure are generated from the average result of 31 different independent runs of MOEA/D for each of the 25 combinations.

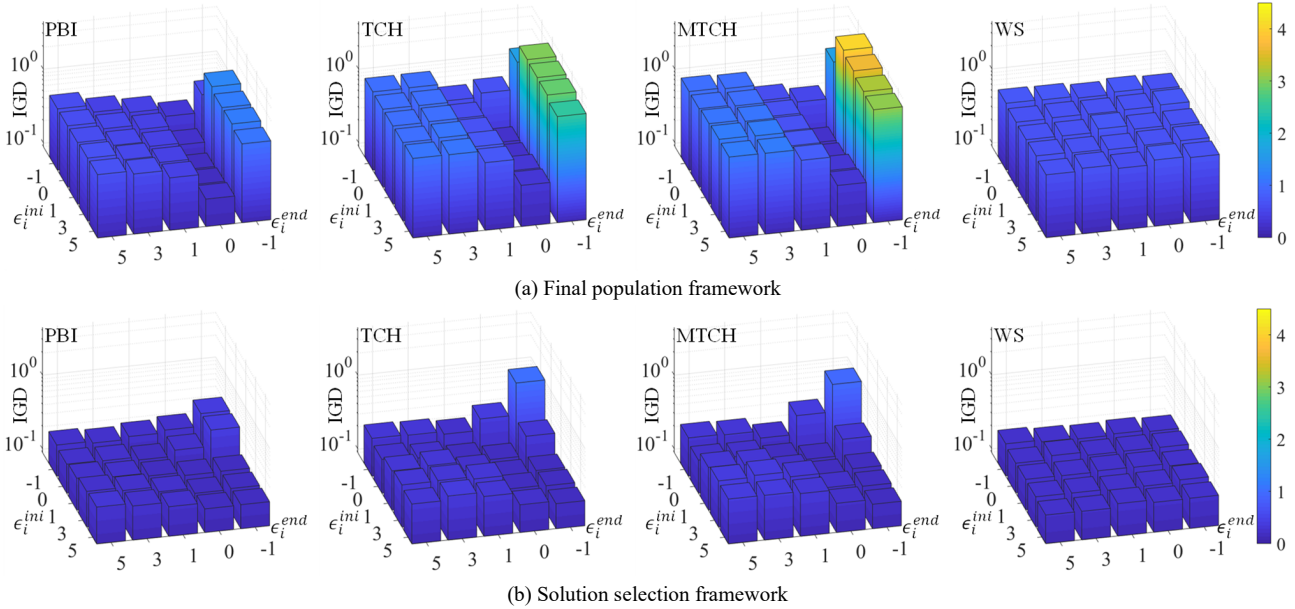

(a) Final population framework

(b) Solution selection framework

Fig. 1. Experimental results by various combinations of initial and final reference point specifications on the three-objective WFG2 problem

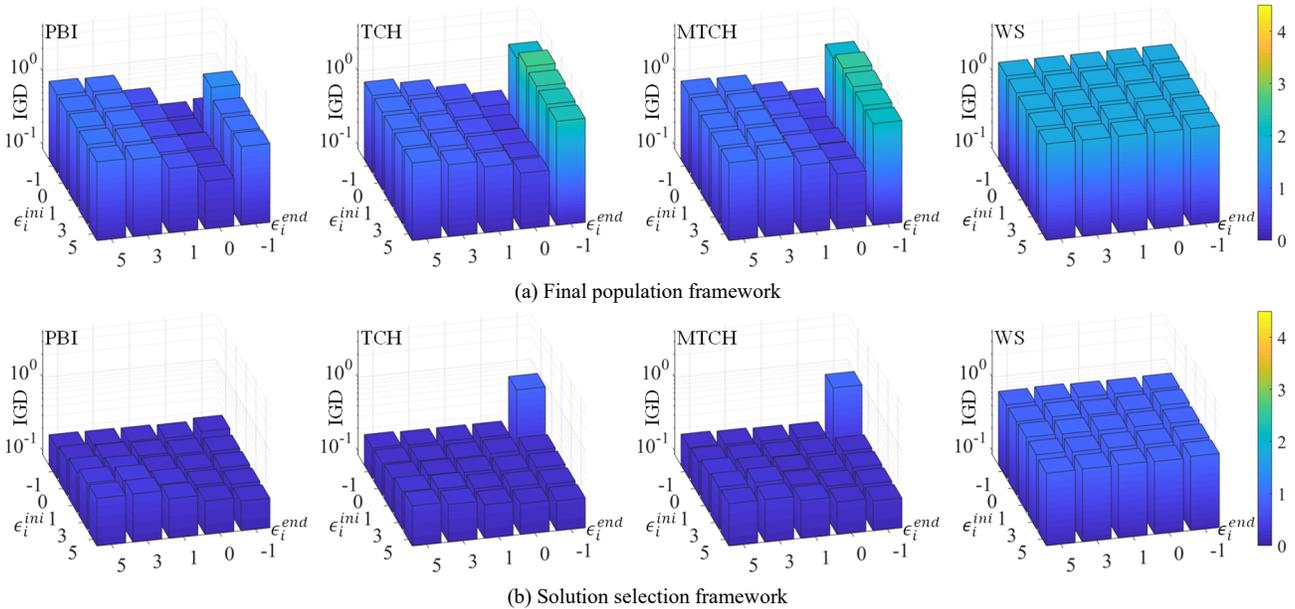

(a) Final population framework

(b) Solution selection framework

Fig.2. Experimental results by various combinations of initial and final reference point specifications on the three objective Minus-WFG2 problem

In each graph of the final population framework in Fig. 1 and Fig. 2, the best (or almost the best) result is obtained from the standard combination $(0, 0)$ (i.e., $(\epsilon_i^{ini}, \epsilon_i^{end})$). That is, we can say that the use of the estimated ideal point for the reference point is appropriate in the final population framework. For the solution selection framework, better performance of MOEA/D is obtained from other combinations. For example, in Fig. 2(b), better results were obtained from $(5, -1)$ than $(0, 0)$: the average IGD value by MOEA/D-TCH is 0.2316 with $(0, 0)$ and 0.2298 with $(5, -1)$. It is clear from Fig. 1 and Fig. 2 that better results were obtained from the solution selection framework in (b) than the final population framework in (a).

We can also see that the results from the solution selection framework are robust over a wide range of settings except for the combination of $(-1, -1)$. The combination $(-1, -1)$ is not a good choice especially for the Tchebycheff and modified Tchebycheff scalarizing functions. When the combination $(-1, -1)$ is used, $\epsilon_i$ is always equal to $-1$. This means that the reference point is the estimated nadir point in the normalized objective space, which is clearly an inappropriate setting. This is the reason why the performance of MOEA/D is bad when the final setting is $-1$ in the final population framework in Fig. 1(a) and Fig. 2(a) independent of the initial setting. However, in the solution selection framework, good results are obtained even when the final setting is $-1$ (if the initial setting is not $-1$). For the other test problems, similar results are obtained.

In summary, the performance of the final population framework is sensitive to the reference point specifications, especially the location of the reference point at the final generation. In almost all cases, the best results in the final population framework are obtained when the estimated ideal point is used as the final reference point (e.g., $\epsilon_i^{end} = 0$ in Fig. 1(a) and Fig. 2(a)). In the solution selection framework, the performance is robust over a wide range of reference point settings. The best results are obtained from other settings than the estimated ideal point (i.e., (0, 0) in Fig. 1(b) and Fig. 2(b)). Moreover, the performance of the solution selection framework over a wide range of settings of the reference point is better than that of the final population framework with its best setting in many test problems (e.g., see Fig. 2).

*B. Genetic Algorithm-Based Hyper-Heuristics*

In this section, a genetic algorithm (GA) is used as an offline hyper-heuristic method to further examine the robustness and flexibility of the solution selection framework for the design of MOEA/D. We use a binary-coded genetic algorithm as the tuner to find the best configuration of MOEA/D in each framework. The GA-based tuner tries to find the best settings from a set of scalarizing functions (i.e., $g \in \{g^{WS}, g^{TCH}, g^{MTCH}, g^{PBI}\}$) and a set of reference point values (i.e., $\epsilon_i^{ini}, \epsilon_i^{end} \in \{-1, 0, 1, 3\}$). All the other parameters in MOEA/D follow the same settings in Section III.A.

The following parameter settings are used for the GA implementation:
Coding: 6-bit binary string,
Population size: 30,
Termination condition: 50 generations,
Crossover: Uniform crossover with the probability 1,
Mutation: Bit-flip mutation with the probability 0.1,
Selection: Tournament selection with tournament size 3,
Fitness evaluation: Average IGD value.

From these settings, we can see that 1500 ($30 \times 50 = 1500$) combinations are examined during the execution of our GA-based hyper-heuristic method whereas the total number of possible combinations is only 64 ($4 \times 4 \times 4 = 64$). We use these unusual settings in order to handle the difficulty of the best algorithm configuration search due to the stochastic nature of multi-objective evolutionary algorithms (i.e., a different solution set is obtained from each run) and performance evaluation (i.e., the reference points for the IGD indicator are generated by obtained solutions). That is, our intention is to choose a near-optimal combination by the GA-based tuner with large computation load and to compare the GA-based tuner results with the results in Section III.A.

The procedure for the GA-based tuner is briefly explained in the following. A population of 30 individuals is randomly generated at the initialization stage. Each individual represents a parameter vector (i.e., a scalarizing function, an initial reference point and a final reference point) for MOEA/D. This means that there are 30 MOEA/D algorithms generated as an initial population. Each MOEA/D algorithm is applied to a test problem and evaluated using the IGD indicator. The reference point set for the IGD calculation is obtained from non-dominated solutions among the obtained solution sets by all the MOEA/D algorithms in the current generation. Thus, the reference point set varies from generation to generation. Each MOEA/D algorithm is evaluated for five times and the average IGD value is calculated. The multiple evaluations of MOEA/D on a test problem are needed to counteract the stochastic effect of evolutionary algorithms. The $(\mu + \lambda)$ generation update mechanism is used, where $\mu$ is the number of parents and $\lambda$ is the number of offspring. At each generation, 30 offspring are generated by genetic operators. Then, 30 best individuals are selected from the combined population of parents and offspring at the current generation. Since the reference point set varies from generation to generation, the $\mu$ parents are re-evaluated using the reference point set of the current generation. The selected 30 best individuals are the parents for the next generation. The process is repeated until the termination condition is met.

The experimental results of the hyper-heuristic MOEA/D are summarized in Table I. For each framework, the GA-based tuner suggests an optimal choice of the scalarizing function and the reference point specifications for each test problem. For each test problem, we performed 31 independent runs using the suggested MOEA/D configuration for each framework. The Wilcoxon's rank sum test at a significant level of 5% is used to evaluate whether there is a significant difference in the performance of MOEA/D between the two frameworks. The symbol '+', '−' or '=' means that the results obtained by the solution selection framework is significantly better than, worse than, or similar to the final population framework, respectively. In general, the search for the best configuration of MOEA/D using a GA-based tuner is not easy. Since each individual is examined by a small number of runs (e.g., each individual is examined by five runs in our experiments), the performance evaluation result of each individual is noisy due to the stochastic nature of MOEA/D. We have carefully compared the experimental results by the GA-based tuner with the results in Section III.A. From the careful comparison, we can see that near-optimal configurations are found by the GA-based tuner in almost all cases.

Table I shows that the solution selection framework can always obtain a better performance MOEA/D configuration than the final population framework. The robustness of the solution selection framework offers a higher flexibility for the design of MOEA/D. Since MOEA/D with the solution selection framework is less sensitive to the parameter settings, a wider range of parameter values can be used to design a good performance algorithm. This can be seen in the columns of Table I for the solution selection framework where the standard reference point specification (0, 0) is not selected and a different setting is selected for a different test problem. On the contrary, the standard reference point specification (0, 0) is often used in the final population framework (i.e., 4 out of 8 test problems). Moreover, when the same scalarizing function and the reference point specification are selected, the performance of MOEA/D with the solution selection framework is higher than that with the final population framework (e.g., WFG2 and WFG3).

TABLE I. EXPERIMENTAL RESULTS OF THE HYPER-HEURISTICS MOEA/D ON THE THREE-OBJECTIVE WFG1-4 AND MINUS-WFG1-4 TEST PROBLEMS. THE STANDARD COMBINATION (0,0) FOR THE REFERENCE POINT SPECIFICATIONS IN MOEA/D IS HIGHLIGHTED USING AN UNDERLINE. THE BEST AVERAGE RESULT FOR EACH TEST PROBLEM IS HIGHLIGHTED USING YELLOW COLOR.

| Test Problems | Final Population Framework | | | | Solution Selection Framework | | | |
|---|---|---|---|---|---|---|---|---|
| | $g$ | $\epsilon_i^{ini}$ | $\epsilon_i^{end}$ | Mean IGD (Std) | $g$ | $\epsilon_i^{ini}$ | $\epsilon_i^{end}$ | Mean IGD (Std) |
| WFG1 | $g^{\text{MTCH}}$ | **0** | **0** | 0.4851 (0.4514) | $g^{\text{PBI}}$ | 1 | 1 | 0.3123 (0.1049) = |
| WFG2 | $g^{\text{PBI}}$ | 1 | 0 | 0.1866 (0.0108) | $g^{\text{PBI}}$ | 1 | 0 | 0.1771 (0.0085) + |
| WFG3[a] | $g^{\text{PBI}}$ | 1 | 0 | 0.1672 (0.0139) | $g^{\text{PBI}}$ | 1 | 0 | 0.1140 (0.0134) + |
| WFG4 | $g^{\text{PBI}}$ | 1 | 0 | 0.2415 (0.0025) | $g^{\text{TCH}}$ | 1 | 0 | 0.2224 (0.0019) + |
| Minus-WFG1 | $g^{\text{TCH}}$ | **0** | **0** | 0.3420 (0.0072) | $g^{\text{TCH}}$ | -1 | 0 | 0.1847 (0.1860) + |
| Minus-WFG2 | $g^{\text{PBI}}$ | **0** | **0** | 0.3503 (0.0040) | $g^{\text{TCH}}$ | 1 | 1 | 0.2428 (0.0047) + |
| Minus-WFG3 | $g^{\text{TCH}}$ | **0** | **0** | 0.2679 (0.0002) | $g^{\text{TCH}}$ | -1 | 1 | 0.1736 (0.0014) + |
| Minus-WFG4 | $g^{\text{WS}}$ | 0 | -1 | 0.2208 (0.0002) | $g^{\text{WS}}$ | 3 | 1 | 0.2193 (0.0021) + |

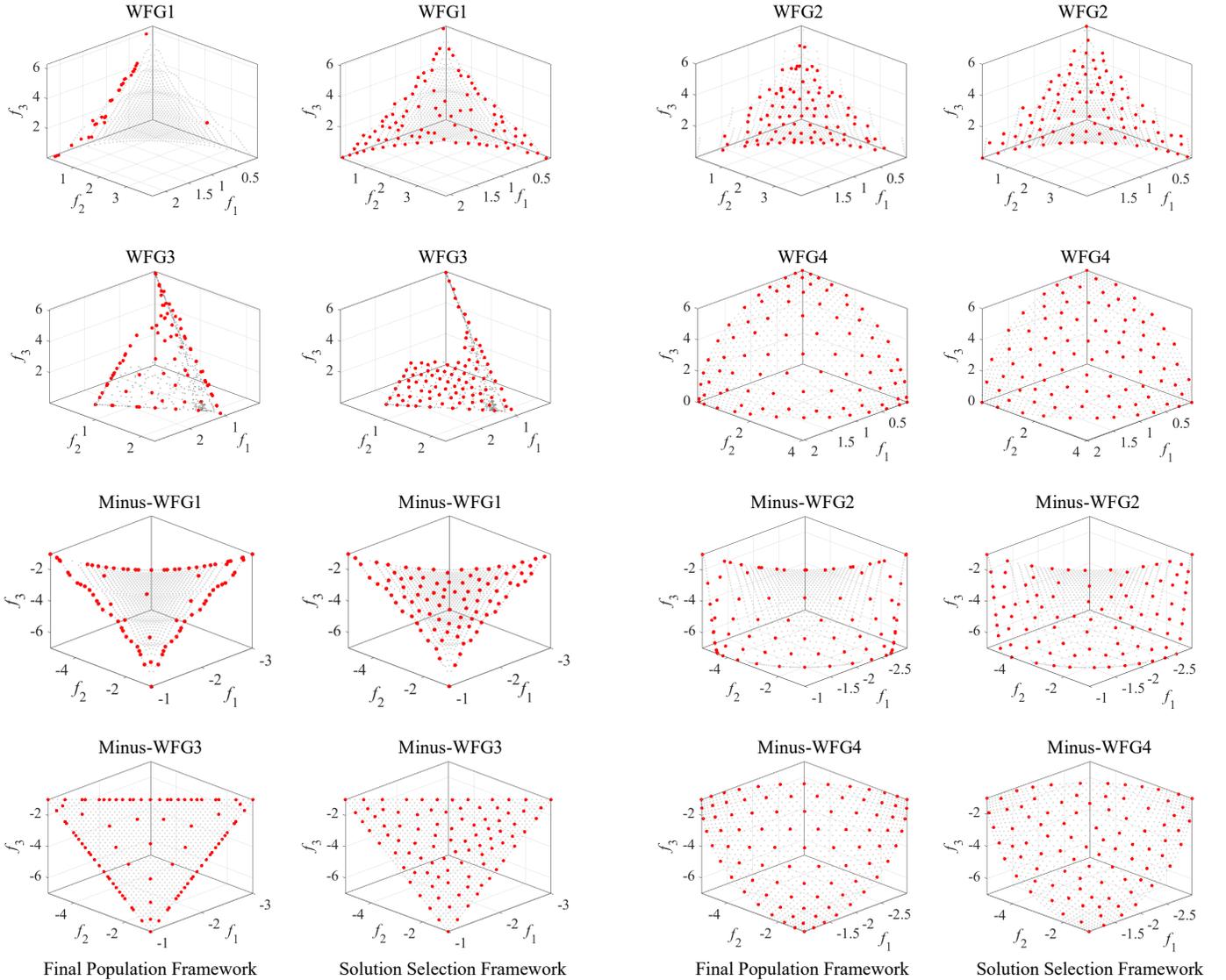

Fig. 3: Experimental results of the selected median run from 31 independent runs of the obtained MOEA/D configuration on each test problem under the final population and solution selection frameworks. The red points are the solutions obtained by MOEA/D algorithms using the parameter settings suggested by the GA-based hyper-heuristic method in Table I for each framework. The gray points are the reference points (obtained from the PlatEMO) for each test problem.

---

[a] We checked the IGD reference points for WFG3 in PlatEMO, and found that they do not cover the flag region. We also checked the IGD reference points in JMetal. They also cover the line part only. Thus, we generated a reference point set by choosing non-dominated solutions from solution sets obtained by different EMO algorithms. We used NSGA-II, NSGA-III, MOEA/D-PBI, SMS-EMOA and SPEA2 with the population size 100 and 100,000 solution evaluations. A total of 406 non-dominated solutions were obtained, which were used as IGD reference points for WFG3.

Another advantage of the solution selection framework is that a set of more uniformly distributed solutions can be obtained from the unbounded external archive. This is illustrated in Fig. 3 where the obtained solution set by a representative single run of the designed MOEA/D (with the median IGD value among 31 runs) is shown for each framework. It is clearly demonstrated that well-distributed solutions are obtained from the solution selection framework.

For Minus-WFG4, the best performance is obtained in Table I for the two frameworks when the weighted sum function is used. As we have mentioned in Section III.A, the search performance of the weighted sum function does not depend on the reference point specification. Thus, for Minus-WFG4, the selected reference point specifications have no meaning.

*C. Discussions*

In this subsection, we discuss the solution selection methods in the solution selection framework. The solution selection method is important in the solution selection framework. In this paper, the distance-based selection method is used in our study because it is fast even for a large solution set. In general, the distance-based selection method is an efficient way to select a set of solutions from an unbounded external archive (since the unbounded external archive usually contains a large number of non-dominated solutions). As shown in Table I, the distance-based selection method works well in all cases. Better results are obtained from the solution selection framework with the distance-based selection method than the final population framework. However, one potential issue with the distance-based solution selection method is that the quality of the selected solutions is not necessarily good. This is because the distance-based selection method does not directly optimize any performance indicator.

Apart from the use of distance-based selection methods, performance indicators such as the IGD and hypervolume indicators can be used to select a set of solutions in the solution selection framework. The solution selection method using a performance indicator is not new [23]. In fact, it has been used as a postprocessing procedure to select a set of solutions from an archive for performance evaluation/comparison. One issue with the IGD-based selection method is the reference point set. A reference point set is needed for the IGD calculation. It is difficult to obtain a reference point set in real-world problems since the Pareto fronts are often unknown. In this situation, it may be possible to use the obtained non-dominated solutions in the unbounded external archive as the reference point set if one wants to optimize the IGD indicator value. In summary, an indicator-based selection method is useful when the indicator value of the final solutions needs to be optimized. A potential disadvantage of using an indicator-based selection method is that a longer computation time may be needed for the solution selection from a large archive. However, this disadvantage will be overcome with the advancement of the computing power and the development of efficient calculation methods of indicator values (or contribution values).

## IV. CONCLUSIONS

In this paper, we demonstrated that MOEA/D with the solution selection framework has more robust and higher performance than the existing final population framework. Configurations of MOEA/D with high performance can be easily obtained by incorporating the solution selection framework into the algorithm design process. A GA-based hyper-heuristic method was used to show the flexibility and high performance of the solution selection framework in the algorithm configuration of MOEA/D. Our experimental results obtained from the solution selection framework are very encouraging. Thus, the use of the solution selection framework in the design of new high-performance EMO algorithms is strongly recommended.

As a future study, the solution selection framework can be further extended to other EMO algorithms. Since solution selection from the unbounded external archive plays an essential role in the solution selection framework, it is an important future research topic to develop a new efficient and effective solution selection method.